\documentclass[conference]{IEEEtran}
\ifCLASSINFOpdf
  \usepackage[pdftex]{graphicx}
  \graphicspath{ {pictures/}{../pdf/}{../jpeg/} }
  \DeclareGraphicsExtensions{.pdf,.jpeg,.png,.PNG}
\else
\fi

\ifCLASSOPTIONcompsoc
 \usepackage[caption=false,font=normalsize,labelfont=sf,textfont=sf]{subfig}
\else
 \usepackage[caption=false,font=footnotesize]{subfig}
\fi

\usepackage{flushend}
\usepackage{url}
\usepackage[table]{xcolor}
\usepackage{tabularx,booktabs,textcomp}
\newcolumntype{C}{>{\centering\arraybackslash}X} 
\newcolumntype{L}{>{\raggedright\arraybackslash}X} 
\usepackage[noadjust]{cite}
\usepackage{hhline}

\renewcommand{\thesection}{\arabic{section}}
\renewcommand{\thetable}{\arabic{table}}
\usepackage{soul}
\usepackage{xcolor}
\usepackage{multirow}
\usepackage[vlined,ruled,linesnumbered]{algorithm2e}
\usepackage{wasysym}
\usepackage{lipsum} 

\begin{document}

\title{Evaluating a Novel Neuroevolution and Neural Architecture Search System}

\author{\IEEEauthorblockN{Benjamin David Winter\IEEEauthorrefmark{1},
William John Teahan\IEEEauthorrefmark{2}}
\IEEEauthorblockA{School Of Computer Science and Electronic Engineering\\
Bangor University,
Wales\\
Email: \IEEEauthorrefmark{1}eeu60d@bangor.ac.uk,
\IEEEauthorrefmark{2}w.j.teahan@bangor.ac.uk}}
\maketitle

\maketitle

\begin{abstract}
The choice of neural network features can have a large impact on both the accuracy and speed of the network. Despite the current industry shift towards large transformer models, specialized binary classifiers remain critical for numerous practical applications where computational efficiency and low latency are essential. Neural network features tend to be developed homogeneously, resulting in slower or less accurate networks when testing against multiple datasets. In this paper, we show the effectiveness of Neuvo NAS+ --- a novel Python implementation of an extended Neural Architecture Search (NAS+) which allows the user to optimise the training parameters of a network as well as the network's architecture. We provide an in-depth analysis of the importance of catering a network's architecture to each dataset. We also describe the design of the Neuvo NAS+ system that selects network features on a task-specific basis including network training hyper-parameters such as the number of epochs and batch size. Results show that the Neuvo NAS+ task-specific approach significantly outperforms several machine learning approaches such as Naive Bayes, C4.5, Support Vector Machine and a standard Artificial Neural Network for solving a range of binary classification problems in terms of accuracy. Our experiments demonstrate substantial diversity in evolved network architectures across different datasets, confirming the value of task-specific optimization. Additionally, Neuvo NAS+ outperforms other evolutionary algorithm optimisers in terms of both accuracy and computational efficiency, showing that properly optimized binary classifiers can match or exceed the performance of more complex models while requiring significantly fewer computational resources.
\\
\end{abstract}


\section{Introduction}
As artificial intelligence systems become increasingly embedded in critical applications—from medical diagnostics to autonomous vehicles—the need for efficient, accurate, and specialized neural networks has never been more pressing. While the headlines focus on massive transformer models consuming enormous computational resources, many real-world AI applications demand something different: lightweight, efficient models optimized for specific tasks with minimal latency and energy consumption. In this context, binary classification models remain foundational to practical machine learning deployments across industries.

Artificial Neural Networks (ANNs), loosely modelled after the human brain's neural structure, have become ubiquitous in data mining, optimization strategies, trend identification, and forecasting across virtually every industry \cite{heikkonen,tamgale,scarborough,miller,kumar}. Their theoretical capacity to handle any computational problem solvable by Turing machines \cite{siegelmann} makes them powerful tools for tackling complex classification challenges. However, this power comes with significant challenges in practical implementation.

\begin{figure}[h!]
\centering
\includegraphics[width=1.01\columnwidth]{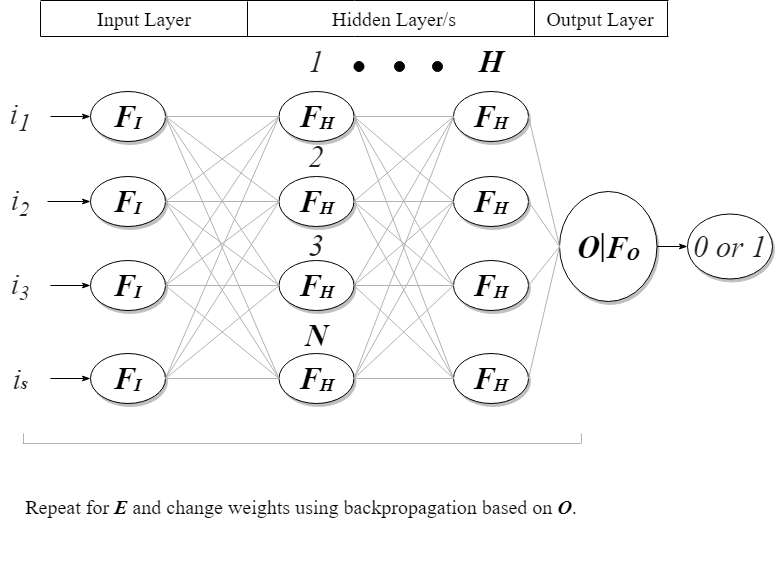} 
\caption{An example of a neural network to solve a binary classification problem. Here $s$ = the size of the input data, $H$ = the number of hidden layers, $N$ = the number of nodes per hidden layer, $F_I$ = the activation function for the input layer, $F_H$ = the activation function for all hidden layers, $F_O$ = the activation function for the output layer, $O$ = the optimiser function, $E$ = the number of epochs.}
\label{fig:network}
\end{figure}

One critical limitation of ANNs lies in their tendency to become trapped in local optima during training, leading to suboptimal performance despite extensive computational investment. Evolutionary Algorithms (EAs) offer a compelling solution by evolving features within neural networks—a technique known as Neuroevolution. This approach enables radical changes to network structure, topology, and other features, allowing the network to escape local optima that would otherwise limit its performance. The neural network topology illustrated in Figure \ref{fig:network} demonstrates the various structural elements that can be evolved to optimize an ANN for a specific classification task.

While recent industry trends have shifted toward large language models and transformer architectures, binary classification models and specialized neural networks remain essential for numerous practical applications. These models offer critical advantages in computational efficiency, interpretability, and resource utilization—particularly important for edge computing, real-time systems, and environments with energy constraints. The optimization of such specialized models therefore continues to be of paramount importance in practical AI deployments.

ANNs tend to be developed homogeneously, for example, using the same activation function throughout the network. The activation function in a neural network is responsible for taking the input from a summing function and turning that arbitrary number into a number that can be read as a positive or negative result for the given task. 

In recent years however, research on activation functions has shifted towards task-specific activation functions. These are functions that are adapted using exhaustive \cite{ramachandran} or evolutionary methods \cite{bingham} based on the results of how well the network computes a specific task. Activation functions are used to add non-linearity to the output of a node. However, traditionally the developer of the neural network must know beforehand whether their network requires non-linearity and foremost, exactly which activation function should be chosen for each layer to provide the best results.

This paper addresses a critical challenge in neural network optimization: how can we systematically determine the ideal architecture and hyperparameters for a specific classification task without exhaustive manual tuning? We present Neuvo NAS+, a novel Neuroevolution and Neural Architecture Search system that automatically evolves optimal network configurations. Our approach differs fundamentally from NEAT \cite{stanley} by focusing not only on topology optimization but also on training hyperparameters such as epoch count and batch size, delivering improvements in both accuracy and training efficiency.

The Neuvo NAS+ system involves the creation of neural network solutions which are maintained in a population and tested to find the optimum results for various datasets. The fitness of each network are assessed on their accuracy each generation, whilst the genetic operations selection, crossover and mutation are applied as standard for an evolutionary approach. The fittest network configuration is returned at the end of the evolutionary run, specifying the network's optimum topology, structure and which network functions were found to be the fittest for each task.

Our experiments demonstrate that properly optimized binary classifiers evolved through our approach can match or exceed the performance of more complex models while requiring significantly fewer computational resources. This efficiency becomes particularly valuable as AI deployment extends to edge devices and applications requiring low-latency predictions that larger models cannot efficiently deliver.

This paper is organised as follows. The related work section provides an in depth analysis of the components of a neural network's architecture and how Neuroevolution can optimise them. This is followed by a discussion of the history of Neuroevolution as well as examining current strategies to overcome problems with traditional ANNs. The methodology and creation of the Neuvo NAS+ system is presented in section 3, with the discussion, results and analysis compared to other binary classification methods provided in section 4 in relation to various fitness metrics. Lastly, section 5 concludes this paper and examines potential future directions for Neuroevolution.

\section{Background and Related Work}
\label{sec:bg}

This section contains an in-depth summary of  neural network features, specifically those that are interchangeable and how they change amongst datasets. An in depth analysis of some Neuroevolutionary techniques follows.

\subsection{The number of hidden layers}
Hidden layers are used in a neural network to provide the capacity to solve non-linearly separable problems. A non-linearly separable problem is one that cannot be solved by classifying the data into two groups.

However, a further question arises when data is known to be non-linearly separable such as "How many hidden layers should be used?" The answer is entirely dependant on the dataset. If the dataset is large and the goal is to train the network to recognise patterns in the data in a form of automatic feature engineering, two or more hidden layers could be beneficial. Other difficulties can also arise in choosing the number of hidden layers. For example, a convolutional neural network with many hidden layers can find patterns between pixel sparsity, pixel colour and pixel density to identify the subject of an image, but there have been cases where changing one pixel on an already trained network can produce very odd results \cite{sui}. This is because patterns have already been established within the hidden layers of the network and any small change from the normal dataset such as a rogue pixel, can completely disrupt the equation of how each pattern is formed and interpreted.

\subsection{The number of neurons per hidden layer}
\label{sec:numNeuron}

A network's number of hidden layers and the number of neurons per hidden layer tend to be developed systematically, with the same number of neurons for every hidden layer. This results in the number of neurons for every hidden layer within the network being dependant on the number of neurons each hidden layer multiplied by the number of hidden layers. This means that the relationship between the neurons and hidden layers greatly dictates the time it takes to both train and test the network. This is purely due to adding more variables to an already significant amount of equations, such as the summing function, activation functions, loss functions and then a loss modifier such as back-propagation.

\subsection{Summing Functions}
\label{sec:sumFunc}
For a fully connected network, the summing function element wise is as follows: 
\begin{align}
\notag
z_1 = (w_1 \cdot x_1) + (w_2 \cdot x_2) + (w_3 \cdot x_3) + b 
\end{align}
where $w_1$, $w_2$ and $w_3$ are the weights  and $x_1$, $x_2$ and $x_3$ are the inputs, and $b$ is the added bias. This calculation is typically vectorised and simplified into the following equation: 
\begin{align}
\notag
z = W \cdot x + b
\end{align}

\noindent where $W$ is a matrix containing all weights of the layer, $x$ is all inputs, $b$ is all biases and $z$ is a vector.

The summing function is used to sum the weights and biases of each neuron in the current layer and create a vector of scalar values to be passed element wise to an activation function.

\subsection{Activation Functions}
\label{sec:actiFunc}
There are many different types of activation functions, all of which can change and modify the output of the network. Activation functions receive a vector from the summing function and perform element wise calculations. For linearly separable problems, activation functions such as ‘Softmax’ can suffice in providing accurate classifications, whereas in a non-linearly separable problem ‘Softmax’ would perform poorly as it is not able to acquire non-linear relationships between hidden layers, which as previously discussed are needed for non-linearly separable data.

In a standard deep neural network, the activation function passes its value on to a loss function which determines the difference between the expected result and the actual result and makes adjustments to the network's weights through the use of back-propagation. Therefore the value from the activation function that is passed on to the loss function is extremely important, as well as being dependant on the dataset in use. This causes a problem if an activation function is chosen for a specific dataset and then applied to a new dataset later on, as the relationships and patterns between the hidden layers are likely to vary with various datasets and make the value from one activation function in one dataset good whilst poor for another.
This step can be considered as being the final step of a feed-forward neural network. All of the data has been forwarded towards an output, but to determine how good an output actually is, we need a function to assess its loss.

\subsection{Loss Function}
\label{sec:lossFunc}
The loss function is used to determine the output error. This is the difference between the predicted value and the actual real known value. In a binary classification problem that has one output, the following equation for binary cross-entropy is commonly used for determining the loss of binary classification problems.
\begin{align}
loss = \sum\limits_{i=1}^1 y_i \cdot log(\hat y_i) + (1 - y_i) \cdot log(1 - \hat y_i)
\end{align}
where $\hat y_i$ is $i$-th scalar value in the model output and $y_i$ is the corresponding target value.

The loss function takes the output vector from the activation function and modifies the weights based on the overall error of the network. For a network with only one output and two attaching neurons, the error of both neurons can be defined by the following equation:
\begin{align}
\notag
E_1 = \frac{w_{11}}{w_{11}+w_{12}} \cdot O
\\
E_2 = \frac{w_{12}}{w_{12}+w_{11}} \cdot O
\end{align}
where $E_1$ is the error of neuron one, $E_2$ is the error of neuron two and $O$ is the output error.

This equation gets extended for networks that have more than one output, for example a network that has two outputs will follow this equation:
\begin{align}
\notag
E_1 = (\frac{w_{11}}{w_{11}+w_{12}} \cdot O_1) + (\frac{w_{21}}{w_{21}+w_{22}} \cdot O_2)
\\
E_2 = (\frac{w_{12}}{w_{12}+w_{11}} \cdot O_1) + (\frac{w_{22}}{w_{22}+w_{21}} \cdot O_2)
\end{align}
where $O_1$ is the output error for output node 1 and $O_2$ is the output error for node 2.

Determining the loss of each neuron is a crucial role is the first step of backpropagation. The weights connecting each neuron can then get changed based on how much ‘error’ its connecting neurons produce. If a dataset requires a large number of neurons or hidden layers, the time used to determine the errors of each neuron can grow rapidly, specifically if the network's architecture is a feed forward network where every neuron is connected by a weight to each neuron in the previous layer.

\subsection{Optimiser functions}
An optimiser function is responsible for tweaking a neuron's weights based on the output of the loss function. However, some optimisers can get stuck in local minima and produce poor results.
Below is an equation to change the weights without optimising against local optima:

\begin{align}
\notag
W = W - l \cdot dW
\end{align}
\begin{align}
\notag
b = b - l \cdot db
\end{align}
where $W$ is the set of weights, $l$ is the learning rate, $dW$ is the derivative of the weights, $b$ is the set of biases and $db$ is the derivative of the biases. It is important to note that these values will change each epoch and are not stored. The learning rate controls how fast the model will adapt to the problem. Small learning rates require more epochs, whereas larger learning rates result in rapid changes and require fewer training epochs but can be unstable and converge on a local minima. The decision to decide what learning rate to use entirely depends on the dataset in use. A simple dataset using a small learning rate would converge slower than one using a large learning rate but not necessarily increase the accuracy.

There are optimizers that try to alleviate the problem of being stuck in local minima; these are called momentum-based optimizers. 
Momentum in the scope of an optimiser function is used to increase or decrease the leap of change based on how large the error is. If the error is small, then the leap of change will be small. On the other hand larger steps are taken if the error rate is large. This is done by taking the exponentially weighted averages of the derivatives:

\begin{align}
\notag
VdW = \beta \cdot VdW + (1 - \beta) \cdot dW
\end{align}
\begin{align}
\notag
Vdb = \beta \cdot Vdb + (1 - \beta) \cdot db
\end{align}
where $\beta$ represents momentum and ranges from $0$ to $1$, typically set to $0.9$ to smooth out the steps of gradient descent. After calculating the averages, the original equations can be updated:

\begin{align}
\notag
W = W - l \cdot VdW
\end{align}
\begin{align}
\notag
b = b - l \cdot Vdb
\end{align}

\subsection{Batch Size}
The batch size represents the number of training data samples that will be passed through the network before the network's parameters are adjusted to the error rate.

There are three techniques used when selecting a batch size: Stochastic Gradient Descent (SGD), Batch Gradient Descent (BGD) and Mini-batch Gradient Descent (MBGD).

SGD sets the batch size to one. Therefore, the network takes only one sample and trains based on the error rate and so on for $n$ samples. This is a problem for large datasets as it results in the training process taking a longer time processing a large number of samples one at a time. In other words, it trains, backpropagates and evaluates its training loss one sample at a time. Moreover, one must consider that the smaller the batch size, the less accurate the estimate of the gradient will be. 

BGD could be thought of as the opposite to SGD where instead of minimising the batch size, it looks to maximise the batch size with respect to the number of samples in the training set. BGD updates the network less times than SGD so it is less computationally expensive. However, it tends to produce premature convergence and gets stuck in local optima due to its inherently unstable error gradient. 
 
MBGD lies in the middle between BGD and SGD and typically sets the batch size to 16, 32, 64 or 128. Mini-Batch Gradient Descent is typically used to find a balance between the accuracy of the gradient and complexity/time.

When setting a batch size, one must consider the size of the dataset, the size of each sample, and the machine in use's capability and memory capacity. Furthermore, if mini-batch gradient descent is used, there is still the factor of determining exactly what batch size to use. This level of fine-tuning accompanied by the fine-tuning of other network parameters quickly becomes unmanageable if doing so manually.

\subsection{Summary}
For each specific solvable task, an optimum neural network architecture exists. However, finding it can be challenging. This is particularly true when factoring in the bias-variance dilemma, which is encountered when trying to minimise error whilst keeping the network applicable beyond the network's training data. The difficulty of finding the optimum network architecture is further enhanced if there is a multi-objective metric being used to assess a network's functionality. 

Neuroevolution can alleviate this dilemma by automatically iterating through a search space of possible network parameters, eventually through evolution finding an optimum solution. This is especially important in networks that take a significant time to train by making sure that any time spent changing the network's hyper-parameters will not be wasted.

\subsection{Neuroevolution}
One of the first approaches to using evolutionary algorithms to optimise neural networks (Neuroevolution) was developed by Montana et al. \cite{montana}. They used a genetic algorithm (GA) to explore a large complex error space to find a global minimum and avoid  local minima. They trained and tested their network using the popular Sonar dataset \cite{sonar} to classify whether a rock or a metal cylinder was present in a water terrain. They hypothesised that typical neural networks that use back-propagation often fail to find the global minima in a short amount of time. Their solution was to use a genetic algorithm as opposed to back-propagation to improve the training of networks, in order to find near globally optimum weights whilst gradually enhancing the complexity of the network features. They found that a genetic algorithm performed better at finding a global optimum in less time than back-propagation. However, the genetic algorithm would struggle if they had continuously changing data as it would require an ever changing stochastic genetic algorithm.

A GA-based Neuroevolution solution commonly uses an encoding of a neural network's weights into a bit-string \cite{holland}. Ronald et al. \cite{ronald94} used a GA to change the structure of their network as well as optimise the network's weights. Traditional neural networks use back-propagation to train the network. However, this often requires large amounts of training data to be accurate. Unsupervised models such as evolutionary algorithms are able to overcome the challenge of having a model with an inherent lack of training data. 

Shenfield et al. \cite{shenfield} discusses a novel multi-objective evolutionary algorithm which optimises the structure, weights and biases of a network as a whole. Each network structure was encoded as a bit-string consisting of the size of each hidden layer, the input layer weights, the hidden layer weights, hidden layer biases and lastly the output layer biases. This bit-string would then evolve, before being decoded for evaluation to ultimately find an optimal network to solve a given classification problem. Their algorithm was found to have slightly worse performance in overall classification accuracy than a standard feed-forward ANN; however it outperformed its rival with respect to minority class recognition.

NeuroEvolution of Augmenting Topologies (NEAT) was developed by Stanley et al. \cite{stanley} and is an example of the combination of a Genetic Algorithm with an ANN. NEAT evolves the topology and weights of an ANN simultaneously based on the fitness of the network's output. NEAT uses a direct encoding scheme which means every connection and neuron is explicitly represented. This version of NEAT starts with a simple perceptron-like neural network consisting of only an input layer and output layer. As the network evolves, the complexity of the network's topology and structure can change, either by inserting a new hidden layer, or changing the amount of nodes per layer. Results show that NEAT outperforms a fixed-topology method named ESP \cite{gomez} when solving a challenging benchmark reinforcement learning task called the Pole-balancing task \cite{pole}. The downfall to starting as a perceptron and incrementally increasing the complexity of the network, is that if the task requires a large complex network architecture, it could take a significant amount of time and computing resources to find the optimum solution. 

Gomez et al. \cite{gomez} devised NNGA which uses a fixed topology in terms of the number of hidden layers but evolved the number of nodes per hidden layer, initial layer weights, biases and $\mu$ coefficient to improve the speed of analysing the aerodynamic databases for various aircrafts. Analysis is used to evaluate an aircraft's stability, control characteristics, flight performance and handling qualities. Results show that NNGA not only improved the speed of analysis over the typically used Navier-Stokes CFD analysis~\cite{navierStokesAnalysis}, but it is also statistically more accurate.

Grammatical Evolution (GE) is an evolutionary algorithm developed by O'Neil et al. \cite{ONeil2003}.
GE uses a Backus-Naur Form grammar to create solutions typically in the form of a bit-string or a char-string, which is then used as a genotype. The genotype is then mapped to a phenotype using the grammar. For example, in the case of Grammatical Neuroevolution, the phenotype is the structure of the neural networks. This makes GE unique in its offering of modularity between genotype and phenotype.

DSGE is a model developed by Assun{\c{c}}{\~{a}}o  et al. \cite{assuncao} that used Grammatical Evolution for Neuroevolution to create dynamic rules that specify the connection possibilities of neurons in the network. DSGE was developed to solve the problem with Grammar-based Genetic Programming (GGP), which is that GGP is limited to the evolution of networks with only one hidden-layer. DSGE allows multiple hidden layers to be added through a recursive mapping process. DSGE also allows parameters such as the number of neurons per hidden layer and the weights of each connection to be placed into genotypes. 

\subsection{Genetic operations}
\noindent \textbf{Mutation} -- The mutation operator serves as a mechanism for evolutionary algorithms to maintain diversity and prevent the population from becoming trapped in local optima. In Neuroevolution, a randomly selected feature within the genotype is altered to a new pseudo-random feature from a predefined search space of available features.

\noindent \textbf{Selection} -- Genotypes are chosen according to their fitness. In Neuroevolution, an individual's fitness is evaluated by comparing the network's predictions or outputs to the actual test results. This means the fitness of a network remains unknown until the genotype has been mapped to its corresponding phenotype, and the resultant network has been trained, compiled, and tested.

Two selection operators are prevalent, Tournament selection~\cite{goldberg1991comparative} and Roulette Wheel selection~\cite{holland1973genetic}. For the first, Tournament selection, typically two genotypes are chosen at random from the population, and whomever is the fittest of the two will be selected for reproduction. This technique is useful for populations where there are clear disparities regarding the fitness of networks. For the second selection operator, Roulette Wheel selection, genotypes take a portion of a `roulette wheel' based on their fitness. A random number on the wheel is chosen and the genotype corresponding to whichever portion on the wheel contains that number is chosen for reproduction. This particular operator is favoured for populations that contain networks with similar fitnesses, specifically when it is unclear which feature of the networks are preferred as each network has a fairer chance of being chosen for reproduction.

\noindent \textbf{Crossover} -- Crossover involves the genetic recombination of two individuals within a population, transferring their information to the offspring. In Neuroevolution, various features of a genotype, such as the number of hidden layers, number of nodes per hidden layer, and activation function, can be selected for crossover. There are multiple types of crossover, including one-point~\cite{holland1992adaptation}, two-point~\cite{de1975analysis}, and whole arithmetic crossover~\cite{michalewicz1994genetic}, each determining the extent and manner in which genetic information is passed from the parents to their offspring.

\noindent \textbf{Elitism} -- Elitism involves copying the fittest genotype(s) into the new population ensuring that they will continue into the next generation. It overcomes the problem that these individuals might be lost if they are not selected for reproduction or their genetic make-up is destroyed due to mutation or crossover occurring. Depending on the task, including neuroevolution, elitism can significantly improve a GA's performance~\cite{deb2001controlled}.

\section{Methodology}
This section describes the design and implementation of Neuvo NAS+ that was developed for evolving neural networks that are task specific according to the data being classified. The overall approach is to use a GA to find the fittest phenotype (a network configuration). A population of genotypes which represent candidate configurations is evolved each generation using standard genetic operations for selection, crossover and mutation. The fitness of each network configuration is calculated using the F-Measure of the network for the specific classification task on test data.

The next three subsections describe the devised system: the genes that make up the genotypes, the genotype to phenotype mapping, and the genetic operations used in the design.

\begin{table}[h!]
\centering
\setlength{\tabcolsep}{2.1pt}
\begin{tabular}{|c|p{3.3cm}|l|}
\multicolumn{1}{l}{\bf Gene} & \multicolumn{1}{l}{\bf Gene description} & \multicolumn{1}{l}{\bf $\in$ Gene set} \\ \hline
$H$ & Num. of hidden layers & \{1..3\} \\ \hline
$N$ & Nodes per hidden layer & \{1..64\} \\ \hline
$F_I$ & Activation functions: & \{{\it Elu, ReLu, Sigmoid,} \\
$F_H$ & Input ($F_I$), Hidden & {\it ~~Softmax, Softplus,} \\
$F_O$ & ($F_H$), Output ($F_O$) & {\it ~~Softsign, Tanh}\} \\ \hline
$O$ & Optimiser &\{{\it Adadelta, Adagrad, Adam,} \\
& & {\it ~~Adamax, Ftrl, Nadam,} \\
& & {\it ~~RMSProp, SGD}\} \\ \hline
$E$ & Number of epochs & \{1..100\} \\ \hline
$B$ & Batch size & \{1..64\} \\ \hline
\end{tabular}
\newline
\caption{The eight genes that comprise the genotype used in the design of Neuvo NAS+ that describe different features which determine a particular network configuration.}
\label{tab:genes}
\end{table}
\subsection{Genotypes}\label{sec:search}
A genotype in our system represents a possible network configuration. It is initially created by randomly generating values for eight `genes' that map to parameters that specify each neural network's structure and topology. The genes consist of various network features such as the number of hidden layers, the number of nodes per hidden layer and so on. The description of the genes used in the genotype is shown in Table \ref{tab:genes}.

Overall, there are $\sim$68.8 million possible network configurations (for example, referring to the description of the gene set in the table, there are 3 possible values for the $H$ gene, 64 values for the $N$ gene, and so on). This makes a brute force search approach to find the optimal configuration unfeasible and mandates the adoption of an evolutionary algorithm instead. 

Our neuroevolutionary model uses a fixed length list in Python to represent the genotype. This is mapped to a function that will define the structure of a Keras~\cite{keras} neural network (the phenotype). Keras is an open source neural network library used in the implementation of Neuvo NAS+. The fixed-length genotype always contains the number of hidden layers, the number of units per layer, activation functions for all layers, optimiser, number of epochs and a batch size. Structurally it was designed so that there will always be at least one input layer, one hidden layer and one output layer. All aspects of the genome are randomly generated initially within their respective bounds as seen in Table~\ref{tab:genes}.

The optimiser function determines how quick the search for minima is, depending on the data. As with an activation function, an optimiser within a network can be configured for specific tasks as well. Some tasks will require a momentum based optimiser whilst other tasks that have a small search space and many different close local minima might prefer a non-momentum based optimiser. For this reason, the type of optimiser function is also included as a gene inside the Neuvo NAS+ genotype. 

The activation functions and optimisers used are from Keras' built-in lists of activation functions and optimisers. There are seven activation functions to choose from as the activation function `exponential' was removed from the search space due to inappropriateness with the chosen datasets. There are also eight optimisers to choose from. Keras activation functions are passed in as strings for each of the different types of layers, whereas whilst the optimisers are also passed in as a string, they are only used when the Keras model is being compiled (see line 7 of Algorithm \ref{alg:create}). The fact that they can be passed in as a string instead of a function makes them ideal for encoding as a gene in the genotype.

All genes in the genotype for each member of the population are initially randomised when the genetic search algorithm starts.
The value for the first gene $H$ is set to a random number between one and three. In a binary classification problem specifically, three hidden layers has been found to learn complex representations in a form of automatic feature engineering between layers and adding more hidden layers may improve the time the network takes to train but may not improve its accuracy in classifying.

The values for the second and last genes (the number of hidden layers $N$ and the batch size $B$) are initially generated as a random number between one and 64.

For the seventh gene (the number of epochs $E$), the values were generated within the range: $N = [1, 100]$. This is to allow comparison with the simple ANN we compare against which has its number of epochs set to 100. The reason is that given a task-specific network architecture with less training epochs, it should still beat a non task-specific architecture in terms of accuracy. 

One of the parameters of a network's structure that is not included in the Neuvo NAS+ genotypes involves the decision of whether a bias should be used per network layer. By default, a bias exists on all layers of the network, bar the output layer of course. A further network feature is the loss function; this is set to `{\it binary crossentropy}' as we are testing on binary classification datasets only. The final network feature that is not included within the genotypes are the input dimensions. These are set before the program is run to the number of training attributes per dataset.

\subsection{Genotype to Phenotype Mapping}
The genotype to phenotype 
mapping used by the system is as follows. The system will pass a genotype's contents into a Keras Sequential model, which will build the network using Dense layers, train it and test it on the chosen dataset. As Neuvo NAS+ presently only solves binary classification problems, an output check is run after the network has been compiled, trained and tested on the data. The check changes the output so that outputs that are $\geq 0.5$ become 1, and predictions that are $< 0.5$ become 0. This is customary when solving binary classification problems as there are only two possible answers, 0 or 1.

This paper provides an experimental evaluation of an implementation within the Neuvo framework~\cite{neuvo} called Neuvo NAS+. Neuvo is a novel Python library to allow developers to quickly and efficiently run a Neural Architecture Search to optimise a neural network for their specific dataset. Neuvo is implemented in Python and built upon Keras. Keras~\cite{keras} is an open-source neural network library in Python. Keras and TensorFlow use an object called a tensor that is created as the summation of the input layer, and passed throughout the network. Keras focuses on modularity which makes it ideal when used in conjunction with modular algorithms such as Neuvo NAS+. Furthermore, Neuvo NAS+ utilises a function that takes the content of a genotype as its input and creates a phenotype by feeding the genotype's parameters into a Keras Sequential model to create neural networks (see Algorithm~\ref{alg:create}). 

The pseudo-code of the creation, training and inference in Keras is found in Algorithm~\ref{alg:create}. 

\begin{algorithm}[htb]
\KwIn{genotype $G$, dataset $D$}
\KwOut{Predictions for the binary dataset $D$.}
$H, N, F_I, F_H, F_O, O, E, B \gets G[0], \ldots, G[7]$ \\
\textit{classifier} $\gets$ create new Sequential model in Keras\\
\textit{classifier}.Create input layer($N$, $F_I$)\\
    \For {\upshape number in $H$}
         {  \textit{classifier}.Add hidden layer($N$, $F_H$)}
\textit{classifier}.Add output layer($1$, $F_O$)\\
\textit{classifier}.Compile($O$)\\
\textit{classifier}.Fit($B$, $E$)\\
\textit{classifier}.Predict results for the chosen dataset $D$\\
convert \textit{classifier's} output to binary (0 or 1)
\caption{The procedure used for training and testing a neural network configuration for our neuroevolutionary model.}
\label{alg:create}
\end{algorithm}

\begin{table*}[h!]
\centering
\begin{tabular}{|p{1.5cm}|p{1.5cm}|p{1.5cm}|p{1.9cm}|p{1.8cm}|p{1.6cm}|p{1.5cm}|p{1.5cm}|p{1cm}|} 
 \hline
 Number of hidden layers ($H$) & Nodes per hidden layer ($N$) & Activation Function: Input ($F_I$) & Activation Function: Hidden ($F_H$) & Activation Function: Output ($F_O$) & Optimiser ($O$) & Epochs ($E$) & Batch size ($B$)\\ [0.5ex]
 \hline
\end{tabular}
\newline
\caption{The format for the genotype used by Neuvo NAS+ consisting of the number of hidden layers, number of nodes per hidden layer, activation function for the input layer, activation function for all hidden layers, activation function for the output layer, optimiser, number of epochs and batch size.}
\label{table:geno}
\end{table*}

\noindent \subsection{Genetic operations} \label{sec:genOp}
\noindent \textbf{Mutation} -- The rate of mutation has been set so that there is a 1\% chance of mutation occurring every generation and will never occur on the fittest network. When a mutation does occur, a gene in the genotype is chosen randomly for mutation. This involves randomly selecting an element from the corresponding gene set for that gene. Once a gene is selected, a random value for the designated gene will replace it inside the genome. For example, if an activation function is chosen to be mutated, another activation will be randomly generated to replace it. This can be seen in Table~\ref{table:mutation}.

\begin{table}[h!]
\centering
\begin{tabular}{|r|r|l|l|l|l|r|r|} 
 \hline
 2 & 16 & {\it ReLu} & {\it Tanh} & {\it Sigmoid} & {\it Adam} & 90 & 16\\ [0.5ex]
 \hline\hline
 2 & 16 & {\it ReLu} & {\it \textcolor{blue}{Elu}} & {\it Sigmoid} & {\it Adam} & 90 & 16\\ [0.5ex]
 \hline
\end{tabular}
\newline
\caption{A random mutation is applied to the fourth gene ({\it Tanh}) that replaces the value in the gene with a random choice ({\it Elu}) of the same feature type, shown in blue.}
\label{table:mutation}
\end{table}

\noindent \textbf{Selection} -- Two selection operators were tested: Tournament selection; and Roulette Wheel selection. The former was found to generate better results for our neuroevolutionary model. Tournament selection takes two genotypes at random from the population and evaluates their fitness based on how accurate the network's guesses were compared to the actual results (according to the F-Measure calculation on test data).

\noindent \textbf{Crossover} -- Once two genotypes are selected, they are passed to a crossover function. Neuvo NAS+ uses one-point crossover to reproduce genotypes as this was found to make the most radical changes per reproduction. A random number $r$ is chosen between 0 and the number of genes in the genotype (7) inclusively. Genes with the index $r$ or less within the genotype are swapped with the genes from the other selected genotype and placed into two offsprings. The parents are then removed from the population and the offspring is placed into the population.

This crossover method gives a 1 in 8 chance that individuals will fully swap  their genetic material and pass them on to their offspring, resulting in offspring having identical genetic information as their parents. This was found to slightly reduce variability and occasionally enhance stability, which depending on the dataset was found to be useful.

\noindent \textbf{Elitism} -- The two fittest individuals from the population are determined to be elite. These elite individuals are networks that have the best classification F-Measure between their guesses and the data's test results. During each generation, if a network within the current population has a higher fitness than the elite network, the elite network becomes the new fittest individual. The elite individuals at the end of the run is copied into the next population.

After any change to an individual via reproduction or mutation, the chosen network's offspring must be retrained and tested. If the network gets 100\% of guesses correct (achieves a fitness of 100\%), the details of the network is outputted and the current run terminated.

\section{Experimental Setup}
This section discusses the datasets used, parameters (seen in Table~\ref{table:2}), libraries and hardware specifications used to create our neuroevolutionary model.

\subsection{Datasets}\label{datasets}
Four binary classification datasets have been selected for the experimental evaluation (see Table~\ref{table:3}). Some pre-processing of the data was needed to modify the output from a string to an integer. The Sonar dataset for example used an output of `m' for whether the object was a metal or `r' for if the object was a rock; both of these had to be converted to $0$ and $1$ respectively for compatibility with the code. Various datasets were used to compare Neuvo NAS+ with other neuroevolutionary binary classifiers \cite{assuncao}. The fitness of each network is determined by the F-Measure results of its test data.

Each dataset was shuffled for every run, and 5-fold cross-validation was applied. The results shown are the average across all 5 folds.

\begin{table}[h!]
\centering
\resizebox{\columnwidth}{!}{%
\begin{tabular}{|l|r|r|l|} \hline
 \textbf{Name} & \textbf{Attrib.} & \textbf{Instances} & \textbf{Classification} \\ [0.5ex] \hline\hline
 Heart~\cite{heart} & 14 & 303 & Heart Disease Prediction\\ [0.5ex] \hline
 Pima~\cite{pima} & 9 & 768 & Diabetes detection\\ [0.5ex] \hline
 Sonar~\cite{sonar} & 60 & 208 & Mine or Rock\\ [0.5ex] \hline
 WBCD~\cite{breastcancer} & 32 & 569 & Tumours\\ [0.5ex] \hline
\end{tabular}%
}
\newline
\caption{Datasets used to evaluate Neuvo NAS+, ordered in ascending order by size (attributes $\times$ instances).}
\label{table:3}
\end{table}

\begin{table}[h!]
{\small
\setlength{\tabcolsep}{2.2pt}
\centering
\begin{tabular}{|l|r|l|r|l|r|}
\multicolumn{2}{l}{\bf GA parameters:} & \multicolumn{2}{l}{\bf GA parameters:} & \multicolumn{2}{l}{\bf NN parameters:} \\  \hline
Population size & 25 & Mutation rate & 0.1\% & Hidden layers & $\leq$3\\ [0.5ex] \hline
Generations & 200 & Tournament size & 2 & Batch size & $\leq$64\\ [0.5ex] \hline
Crossover rate & 77\% & Elitism size & 2 & Epochs & $\leq$50\\ [0.5ex] \hline
\end{tabular}
\newline
\caption{Experimental GA and NN parameters.}
\label{table:2}
}
\end{table}

\section{Experimental Results and Discussion}
This section discusses the results from the experimental evaluation as shown in Tables~\ref{table:6} to~\ref{table:5}.

Table~\ref{table:6} shows the Mean Absolute Error, Root Mean Square Error, and F-Measure for the performance of Neuvo NAS+ on the five datasets detailed in Table~\ref{table:3}.

\begin{table}[h!]
{\small
\centering
\hskip0.5cm
\begin{tabular}{|l|r|r|r|} 
 \hline
 \textbf{Dataset} & \textbf{MAE} & \textbf{RMSE} & \textbf{F-Measure}\\ [0.5ex] 
 \hline\hline
 Heart & 0.358 & 0.413 & 0.566\\ [0.5ex] \hline
 Pima & 0.359 & 0.422 & 0.656\\ [0.5ex] \hline
 Sonar & 0.162 & 0.264 & 0.729\\ [0.5ex] \hline
 WBCD & 0.151 & 0.236 & 0.920\\ [0.5ex] \hline
\end{tabular}
\newline
\caption{Neuvo NAS+ classification results on the chosen datasets, after 200 generations.}
\label{table:6}
}
\end{table}

\begin{table*}[h!]
{\small
\centering
\begin{tabular}{|l|r|r|l|l|l|l|l|l|r|r|}
 \hline
 \textbf{Dataset} & $H$ & $N$ & $F_I$ & $F_{H1}$ & $F_{H2}$ & $F_{H3}$ & $F_{O}$ & $O$ & $E$ & $B$\\ [0.5ex]  \hline\hline
 Heart & 1 & 24 & $Softmax$ & $Softsign$ & $--$ & $--$ & $Sigmoid$ & RMSProp & 60 & 3\\ [0.5ex] \hline
 Pima & 3 & 26 & $Elu$ & $ReLU$ & $Elu$ & $ReLU$ & $Sigmoid$ & Adam & 48 & 3\\ [0.5ex] \hline
 Sonar & 2 & 70 & $ReLU$ & $Softplus$ & $ReLU$ & $--$ & $Sigmoid$ & Adam & 44 & 2\\ [0.5ex] \hline
 WBCD & 3 & 43 & $Softsign$ & $SeLU$ & $Softmax$ & $Elu$ & $Tanh$ & Adamax & 59 & 4\\ [0.5ex] \hline
\end{tabular}
\newline
\caption{The neural network parameters for each gene for the fittest genomes found for each dataset.}
\label{table:fittest}
}
\end{table*}
Table~\ref{table:fittest} shows the neural network parameters for each gene in the fittest genomes found for each dataset.
For example, the network that achieved the highest fitness classifying the Heart dataset is defined as follows: 1 hidden layer, 24 nodes per hidden layer, {\it Softmax} as the input layer activation function, {\it Softsign} as the activation function for the first hidden layer, {\it Sigmoid} as the output layer activation function and {\it RMSProp} as the optimiser. The network ran for 60 epochs with a batch size of 3.

As is evident from the table, the Neuvo NAS+ algorithm has chosen a wide range of values for each gene specifically tailored for each dataset. There is no consistent pattern from candidate solutions across datasets, except for the dominance of the {\it Sigmoid} output layer activation function being preferred for 75\% of the datasets. In contrast, the parameters $H$, $N$ and $E$ have selected values across the full range, from as low as a single hidden layer ($H = 1$) for the Heart dataset, and $3$ for the Pima and WBCD dataset, for example. These results illustrate the merit of choosing task-specific parameters when designing neural network-based solutions.

Interestingly, for the network architecture that was evolved for the Heart Disease Detection results seen in Table~\ref{table:fittest}. This was the only one that used the optimiser {\it RMSProp} and only one hidden layer. Nevertheless, Neuvo NAS+ still outperformed other machine learning algorithms at classifying this dataset as can be seen in Table~\ref{table:5}.

The fittest network used to classify the Pima dataset consisted of 3 hidden layers and 26 nodes per hidden layer. The activation functions used for the input layer and hidden layers was {\it Elu} and {\it ReLU} whilst the activation function for the output layer was {\it Sigmoid} respectively. The fittest network also used the {\it Adam} optimiser with a batch size of 3 and it ran for 48 epochs.

For comparison to the Neuvo NAS+ results, several algorithms were implemented using scikit-learn~\cite{scikit-learn}: Gaussian Naive Bayes (GNB), decision tree classifier (C4.5) and Support Vector Machine (SVM). All parameters of each were set to default. 

Lastly, an Artificial Neural Network (ANN) was created for comparison and had an architecture as follows: the number of hidden layers = 1; the number of nodes per hidden layer = 16; \textit{Relu} was used for the input and hidden layer activation function; \textit{Sigmoid} was used for the output layer; \textit{Adam} was used as the optimiser; and the ANN trained for $50$ epochs with a batch size of $4$. These results are included in Table~\ref{table:5}. 

\begin{table}[h!]
\centering
\begin{tabular}{|l|l|r|r|r|r|}  \hline
 \textbf{Algorithm} & \textbf{Metrics} & \textbf{Pima} & \textbf{Heart}
 & \textbf{Sonar} & \textbf{WBCD}\\ [0.5ex]  \hline\hline
 \multirow{4}{*}{GNB} & MAE & 0.108 & 0.066 & 0.141 & 0.023\\ [0.5ex]\cline{2-6}
   & RMSE & 0.232 & 0.182 & 0.265 & 0.107\\ [0.5ex]\cline{2-6}
   & F1 & 0.336 & 0.442 & 0.351 & 0.468\\ \hline 
 \multirow{4}{*}{C4.5} & MAE & 0.129 & 0.089 & 0.107 & 0.019\\ [0.5ex]\cline{2-6}
   & RMSE & 0.254 & 0.211 & 0.231 & 0.098\\ [0.5ex]\cline{2-6}
   & F1 & 0.316 & 0.418 & 0.394 & 0.473\\ [0.5ex]\hline
 \multirow{4}{*}{SVM} & MAE & 0.115 & 0.169 & 0.102 & 0.035\\ [0.5ex]\cline{2-6}
   & RMSE & 0.239 & 0.291 & 0.226 & 0.133\\ [0.5ex]\cline{2-6}
   & F1 & 0.298 & 0.372 & 0.416 & 0.450\\ [0.5ex]\hline
 \multirow{4}{*}{KNN} & MAE & 0.132 & 0.162 & 0.117 & 0.028\\ [0.5ex]\cline{2-6}
   & RMSE & 0.256 & 0.285 & 0.236 & 0.119\\ [0.5ex]\cline{2-6}
   & F1 & 0.300 & 0.355 & 0.408 & 0.462\\ [0.5ex]\hline
 \multirow{4}{*}{ANN} & MAE & 0.443 & 0.473 & 0.336 & 0.211\\ [0.5ex]\cline{2-6}
   & RMSE & 0.472 & 0.489 & 0.396 & 0.278\\ [0.5ex]\cline{2-6}
   & F1 & 0.494 & 0.527 & 0.654 & 0.908\\ [0.5ex]\hline
 \multirow{4}{*}{Neuvo NAS+} & MAE & 0.359 & 0.358 & 0.162 & 0.151 \\ [0.5ex]\cline{2-6}
   & RMSE & 0.422 & 0.413 & 0.264 & 0.236 \\ [0.5ex]\cline{2-6}
   & F1 & \textbf{0.656} & \textbf{0.566} & \textbf{0.729} & \textbf{0.920}\\ [0.5ex]\hline
\end{tabular}%
\caption{Comparison of Neuvo NAS+ against various other techniques for classifying a variety of binary classification datasets. The individual who achieved the highest F1 score for each dataset is shown in bold font.}
\label{table:5}
\end{table}

As seen in Table~\ref{table:5}, Neuvo NAS+ model significantly outperforms C4.5, SVM and this basic ANN architecture in classifying the collection of datasets presented.

\begin{figure}[h!]
    \centering
    \includegraphics[width=0.95\columnwidth]{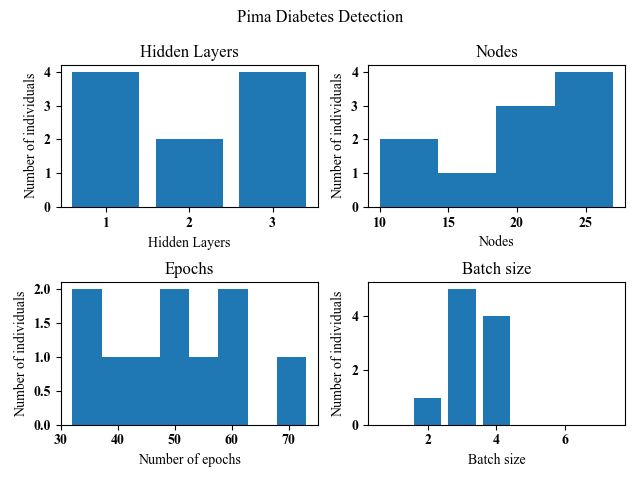}
    \caption{The diversity shown in fit individuals' network architecture classifying the Pima Diabetes Detection dataset.}
    \label{fig:pimaresult_arch}
\end{figure}

\begin{figure}[h!]
    \centering
    \includegraphics[width=0.95\columnwidth]{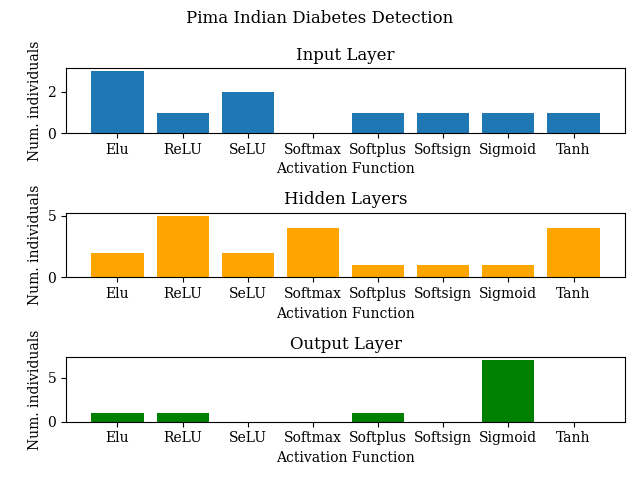}
    \caption{The diversity shown in fit individuals' activation functions classifying the Pima Diabetes Detection dataset.}
    \label{fig:pimaresult}
\end{figure}

Perhaps the most important result is the diversity that is produced amongst genotypes. The genotype diversity for the Pima Diabetes Detection dataset can be seen in Figures~\ref{fig:pimaresult_arch} and \ref{fig:pimaresult}. However, whilst there is marked diversity amongst candidate solutions, there is also an evident convergence on favourable genes such as `Sigmoid' for the output layer activation function.

\subsection{Execution Speed and Improving Performance}
This section discusses the execution speed of our neuroevolutionary model as well as some speed improvements that were implemented.

The execution time in seconds it took NEAT to find an optimum network is shown in Table~\ref{table:speeds}. The procedure used for the creation of the models, and training and testing is specified in Algorithm~\ref{alg:create}.

\begin{table}[h!]
{\small
\centering
\begin{tabular}{|l|r|r|} \hline
\textbf{Dataset} & \textbf{Avg. Training \&}& \textbf{Fittest Training \&}\\ [0.5ex]
 &  \textbf{Testing Speed} & \textbf{Testing Speed}\\ [0.5ex] \hline\hline
Pima & 23.457 & 17.184\\ [0.5ex] \hline
Heart & 3.645 & 3.848\\ [0.5ex] \hline
WBCD & 9.125 & 9.088\\ [0.5ex] \hline
Sonar & 13.581 & 22.984\\ [0.5ex] \hline
\end{tabular}
\newline
\caption{Neuvo NAS+ execution training and testing speeds in seconds.}
\label{table:speeds}
}
\end{table}

In an effort to improve the speed of Neuvo NAS+, the batch size and number of epochs per network were added to each individual's genotype. As previously discussed, if the fittest network only runs for 20 epochs with a batch size of 8, then other networks which are not as fit and have a higher number of epochs and a smaller batch size, are wasting computational time when they are not achieving as good results. To counter this, adding the number of epochs and batch size to the genotype means that these fittest networks can reproduce and create offspring that have a more favourable and efficient number of epochs and batch size respectively. If the population as a whole is not performing well due to a small amount of training epochs, there is a chance that the number of epochs will mutate to a greater number, resulting in more time training the network and potentially fitter networks being added to the population.

Table~\ref{table:speeds} shows the combined training and classification speed. It is assumed that some networks take longer to train because they continue improving training results over a longer period, despite using an early stop callback for training loss.

\section{Conclusion and Future Work}
This paper describes the experimental evaluation of a new system called Neuvo NAS+ that combines Genetic Algorithms, Neuroevolution and task-specific neural network features to solve binary classification problems. Experiments show significantly better results than previous neuroevolution models and other machine learning systems such as a standard ANN, C4.5, SVM, and Na\"{i}ve Bayes \cite{assuncao,ross,hand,cristianini} seen in Table~\ref{table:5}. 

The results illustrate the effective use of task-specific network features such as activation functions, optimisers and number of hidden layers to achieve better and more accurate results. Results also show that a greater number of epochs do not always result in better accuracy. Whilst some results were expected, it is worth noting that the idea of placing network features within an evolving genotype is more feasible than testing every network combination exhaustively, particularly when working on a time sensitive project. 

Future work will evolve more features of the network such as the layer's activation functions as well as determining whether the evolutionary parameters themselves can be included in individual's genotypes, thus evolving every aspect of the network and the optimisation technique in tandem.

\nocite{}
\bibliography{references}
\bibliographystyle{IEEEtran}
\end{document}